# A Corpus for Sentence-Level Subjectivity Detection on English News Articles

**Francesco Antici**[1], **Andrea Galassi**[1], **Federico Ruggeri**[1✉],
**Katerina Korre**[2], **Arianna Muti**[2], **Alessandra Bardi**[2],
**Alice Fedotova**[2], **Alberto Barrón-Cedeño**[2]

[1]DISI, University of Bologna, Bologna, Italy
[2]DIT, University of Bologna, Forlì, Italy
{francesco.antici, a.galassi, federico.ruggeri6}@unibo.it
{aikaterini.korre2, arianna.muti2, alice.fedotova2, a.barron}@unibo.it
alessandra.bardi@studio.unibo.it

## Abstract

We develop novel annotation guidelines for sentence-level subjectivity detection, which are not limited to language-specific cues. We use our guidelines to collect NewsSD-ENG, a corpus of 638 objective and 411 subjective sentences extracted from English news articles on controversial topics. Our corpus paves the way for subjectivity detection in English and across other languages without relying on language-specific tools, such as lexicons or machine translation. We evaluate state-of-the-art multilingual transformer-based models on the task in mono-, multi-, and cross-language settings. For this purpose, we re-annotate an existing Italian corpus. We observe that models trained in the multilingual setting achieve the best performance on the task.

## 1. Introduction

Subjectivity detection (SD) contributes to several Natural Language Processing applications, such as sentiment analysis, bias detection, and fact-checking (Riloff and Wiebe, 2003). Spotting subjectivity is a challenging task, even for human experts. The perception of subjectivity is complex and ambiguous: it may derive from different interpretations of the language, background knowledge, and personal biases (Chaturvedi et al., 2018). Consequently, the creation of corpora for SD is notably difficult and costly (Riloff and Wiebe, 2003).

Standard approaches for corpus creation rely on spotting subjective words (Wiebe and Riloff, 2005; Riloff and Wiebe, 2003) coming from specific lexicons (Das and Sagnika, 2020; Yu and Hatzivassiloglou, 2003; Villena-Román et al., 2015). Nonetheless, these solutions are known to be limited to both domain- and language-specific assumptions (Pang and Lee, 2004a), and require external tools, such as machine translation, for language transfer (Benamara et al., 2011). A few attempts have investigated using annotation guidelines to define general-purpose SD tasks. However, such approaches have demanding issues, including human annotation ambiguity (Chaturvedi et al., 2018), interpretation bias (Geva et al., 2019), and ambiguous case resolution. To minimize the impact of the above issues, we follow a prescriptive paradigm (Rottger et al., 2022) to define the task and the guidelines according to a specific purpose and belief.

In our work, we frame SD into an information-retrieval process, with the purpose of discriminating between sentences from which information can be directly extracted (objective) and sentences that must be further processed (subjective). We propose a data collection methodology tailored to a task-oriented definition of subjectivity. We develop a novel set of annotation guidelines that may be applied to any language, and that can be used to create new SD corpora in other languages. We design an annotation procedure based on the prescriptive paradigm (Rottger et al., 2022). Annotator's subjectivity is discouraged by enforcing a shared belief, while disagreements are mitigated by discussing and resolving controversial cases, resulting in detailed guidelines. We manually curate NewsSD-ENG, a novel high-quality English corpus for SD concerning controversial topics from political affairs in news articles. The corpus contains 1,049 sentences extracted from 23 news articles, out of which 638 end up being objective and 411 subjective. We hope that our corpus can foster the research on subjectivity as a feature for tasks like opinion detection and fact-checking.

We showcase our corpus by employing a set of supervised models for SD, including support vector machines (Cortes and Vapnik, 1995), logistic regressors (Cramer, 2004), transformer-based architectures such as BERT (Devlin et al., 2019) and SBERT (Reimers and Gurevych, 2019). Furthermore, we evaluate if our annotation guidelines can be transferred from one language to another. Since cross-lingual transfer learning has been shown to be effective for low-resource languages with little or no labeled training data (de Vries et al., 2022), we perform multilingual and zero-shot cross-lingual experiments on NewsSD-ENG and SubjectivITA (Antici et al., 2021), an Italian corpus for SD on news articles.

| Profile | Description |
|---|---|
| Gender | 3 identify as male, 4 as female |
| Education | 2 PhD, 4 PhD student, 1 Master student |
| Origin | 5 Central Europe, 2 Eastern Europe |

Table 1: Summary of the annotators' profile.

| Source | No. Articles |
|---|---|
| economist.com | 2 |
| frontpagemag.com | 5 |
| shftplan.com | 6 |
| spectator.co.uk | 2 |
| theguardian.com | 2 |
| theweek.com | 3 |
| tribunemag.co.uk | 5 |
| vdare.com | 3 |

Table 2: Selected British news outlets and the corresponding number of articles samples from these sources.

The employed models achieve on-par or superior classification performance when trained in multilingual settings compared to monolingual ones. Additionally, SBERT performs comparable to its monolingual counterpart when evaluated in the cross-lingual setting. These results suggest that our annotation guidelines can be transferred to other languages.

The contributions of this work are the annotation guidelines, the data collection methodology, and the collected corpus of English news articles.

The remainder of the article is as follows. Section 2 presents our annotation methodology. Section 3 describes our collected corpus. Section 4 and 5 illustrate the experimental setting and discuss results. Section 6 provides background and related work. Section 7 concludes.

## 2. Annotation Guidelines

We recruit seven annotators with near-to-native English command and a linguistics or computing science background. Table 1 summarises the profiles of the annotators, which represent a diverse group of people in terms of gender, (high) level of education, and geographical origin. We define our initial guidelines based on those presented in Antici et al. (2021) and we carry out two pilot annotations to refine them. We manually select articles from eight British news outlets covering controversial political affairs on law, civil rights, economics, and internal politics. Table 2 reports the selected outlets and the number of articles sampled to build our corpus. We select these outlets according to the medium popularity, coverage standpoints (e.g., left vs. right, liberal vs. conservative), and abundance of controversial topics. We choose these topics because they are abundant in opinions and argumentative content (Haddadan et al., 2019). Thus, news outlets covering such topics are likely to use subjective statements in their narratives. Additionally, political affairs are a valuable research playground for fact-checking systems (Graves, 2016).

**First pilot study.** Following our guidelines, an annotator labels a sentence from an article as *subjective* or *objective*. Annotators can only look at the given sentence to perform the annotation. At the end of the study, all annotators discuss ambiguous sentences and analyze edge cases to refine the guidelines. This process aligns with the prescriptive paradigm, where annotator disagreement is a call to action. Furthermore, we refer to linguistic resources (Finegan, 1995; University of Adelaide, 2014) to refine our guidelines concerning edge cases like speculations, biased language, conclusions backed up by factual evidence, and the speaker's emotions. In total, the annotators label 270 sentences from 3 articles. The inter-annotator agreement (IAA), measured as Krippendorf's alpha (Krippendorff, 2011), is 0.40, which can be considered as a fair agreement.

**Second pilot study.** In the first pilot study, we observe that a major point of discussion on edge cases is using contextual information, such as providing sentences adjacent to the one to label. Thus, we carry out the second pilot study to test the impact of considering context when annotating. We instruct half of the annotators to label the sentences using the article context they belong to, whereas the other half annotates as in the first pilot study. In total, the annotators label 70 sentences from 5 articles. The IAA of the group using context is 0.38, while the other achieves a higher IAA of 0.53 ("moderate agreement"). As emerged during the discussions, a possible motivation for this difference is that annotators were more prone to label ambiguous sentences as subjective when the context contained clearly subjective sentences. Besides this gap, annotating using context leads to an increased annotator workload. Thus, annotators label the final corpus by just looking at isolated sentences without considering any context.

**Consolidation.** After the two pilot studies, we consolidate the guidelines to label the final corpus. They are summarized in Figure 1 and reported in Appendix C. The guidelines include a definition of

> A sentence is **subjective** if its content is based on or influenced by personal feelings, tastes, or opinions. Otherwise, the sentence is **objective**.
>
> More precisely, a sentence is subjective if one or more of the following conditions apply:
>
> 1. expresses an explicit personal opinion from the author (e.g., speculations to draw conclusions);
> 2. includes sarcastic or ironic expressions;
> 3. gives exhortations of personal auspices;
> 4. contains discriminating or downgrading expressions;
> 5. contains rhetorical figures that convey the author's opinion.
>
> The following ambiguous cases are **objective**: third-party's opinions, comments that do not draw conclusions and leave open questions, and factual conclusions.
>
> **Note 1:** Reported speech verbatim cannot contain elements that we identify as markers of subjectivity as it is not content created by the writer, and, thus, is **objective**.
>
> **Note 2:** Personal feelings, emotions, or mood of the author, without conveying opinions on the matter, are considered **objective** since the author is the most reliable source for information regarding their own emotions. Emotion-carrying statements are not excluded since they frequently occur in news articles and excluding them from the corpus would turn it less useful in real application scenarios.

Figure 1: Excerpt of our annotation guidelines.

subjectivity, examples, and cases that have been identified as ambiguous during the pilot studies. While Antici et al. (2021) do not propose explicit criteria for addressing quotations, we consider them as third-party opinions and label them as objective. Moreover, we label sentences where the author explicitly states their emotions as objective. This decision may seem counter-intuitive since the presence of emotions is widely used as an indicator of subjectivity (Mihalcea et al., 2012; Veronika, 2006), especially when the purpose is to capture subjectivity as an expression of a "private state" of the author (Riloff and Wiebe, 2003). However, in our work, emotions do not influence the message conveyed in the sentence but rather *are* the content of the sentence. Thus, the sentence is not influenced by emotions.

## 3. Corpus

Inspired by Braun (2023), we perform the annotation process in three stages: (i) two annotators label their assigned sentences; (ii) each couple of annotators discusses ambiguous sentences to reach an agreement; and (iii) in cases where the two annotators fail to agree, a third one labels the disputed sentence. Thus, at least two annotators annotate each sentence in the corpus. The IAA measured on the whole corpus is 0.51 ("moderate") after step (i) and 0.83 ("near-perfect") after step (ii), requiring step (iii) in less than 10% of cases.[1] These IAA agreements suggest that our data collection methodology yields a high-quality corpus for SD.

|       | # Art. | # Sent. | # OBJ    | # SUBJ   |
|-------|--------|---------|----------|----------|
| Train | 16     | 731     | 487 (12) | 244 (46) |
| Dev   | 3      | 99      | 45 (3)   | 54 (8)   |
| Test  | 4      | 219     | 106 (4)  | 113 (16) |
| Total | 23     | 1,049   | 638 (19) | 411 (70) |

Table 3: Corpus statistics. The number of disputed sentences is in parentheses.

Our final corpus, which we name NewsSD-ENG, contains 1,049 sentences, 638 of which are objective and 411 subjective. The statistics of the corpus are summarized in Table 3. Following Cabitza et al. (2023) and Abercrombie et al. (2022), we flag the 89 sentences that required the intervention of a third annotator as a measure of quality assurance (Rottger et al., 2022; Teruel et al., 2018; Davani et al., 2022).

NewsSD-ENG is comparable in size to other sentence- and tweet-level SD corpora (Antici et al., 2021; Bosco et al., 2014; Wiebe et al., 1999a). While there exist English SD corpora with a higher number of sentences than ours (Wiebe et al., 2005), the limited number of sentences in our corpus is symptomatic of our fine-grained annotation methodology, which is oriented to the collection of high-quality data. To demonstrate the task complexity and the need for human expert annotators (Färber et al., 2020), we compare our annotations to those produced by a lexicon-based approach: TEXTBLOB.[2] The average IAA with annotators is -0.21, suggesting that a lexicon-based approach is not suitable for our task.

## 4. Experimental Setting

We formulate SD as a binary classification task where a model has to classify a sentence as subjective (SUBJ) or objective (OBJ). We experiment with two languages to evaluate if our annotation guidelines can be transferred from one language to another to address SD. We consider our English corpus and experiment with Italian as the second language, using the SubjectivITA corpus (Antici et al., 2021). Since SubjectivITA differs from our corpus in how quotes and emotions are labeled (Section 3), we re-annotate those cases by hand to set common annotation criteria between the two corpora. We label the re-annotated corpus as NewsSD-ITA. Further information can be found in Appendix B.

---
[1] We do not measure the agreement after step (iii) since each label assigned in such step would necessarily agree with one annotator and disagree with the other.

[2] https://textblob.readthedocs.io/

We perform a preliminary evaluation considering three settings, following the strategy of Muti and Barrón-Cedeño (2022), where $L_i$ and $L_j$ are each of the two languages involved. In the *monolingual setting* both the training and the testing data are in language $L_i$. In the *multilingual setting* the training data combines both $L_i$ and $L_j$. In the *cross-lingual setting* the training data is in $L_i$ and the test data in $L_j$, in a zero-shot fashion (Huang et al., 2021).

We consider the following classifiers: support vector machine **(SVM)** (Cortes and Vapnik, 1995) and logistic regressor **(LR)** (Cramer, 2004), both with tf-idf representations; multilingual Sentence-BERT **(M-SBERT)** (Reimers and Gurevych, 2019)[3], and MultilingualBERT **(M-BERT)** (Devlin et al., 2019)[4] as in Antici et al. (2021). As baselines, we consider a random uniform **(RND-B)** and a majority **(MAJ-B)** classifier. For evaluation, we report macro F1-score and class-specific F1-scores (F1-OBJ and F1-SUBJ) averaged over three individual seed runs. All models are trained with their default configuration. We fine-tune transformer-based models for 4 epochs following standard practice (Devlin et al., 2019). Additional details are reported in Appendix A.

We aggregate labels via majority voting to train machine learning models (Nguyen et al., 2017). We produce train, development, and test splits such that all the sentences from an article end up in the same split. We balance the number of SUBJ and OBJ sentences in the development and test sets to ensure a sound model evaluation (for NewsSD-ITA, we consider the original splits). Table 3 reports splits statistics.

## 5. Results

Table 4 shows the classification performance on the three settings. In the monolingual setting, M-BERT and M-SBERT greatly outperform the SVM and the LR. The low scores concerning F1-SUBJ achieved by the SVM and LR are possibly due to the unbalanced class distribution in the training set. M-BERT achieves comparable performance to M-SBERT concerning F1-SUBJ, while it outperforms M-SBERT on F1-OBJ.

In the multilingual setting, there is a notable performance improvement compared to the monolingual setting for certain models. For English, M-BERT performs the best, with an improvement of five points on F1-macro over the monolingual setting, followed by M-SBERT with an improvement of two points. Likewise, M-BERT achieves the best performance in Italian, with a three-point F1-macro and a seven-point F1-SUBJ improvement compared to the monolingual setting. In contrast, M-SBERT shows no relevant improvement when trained on either monolingual or multilingual settings.

---
[3]`paraphrase-multilingual-MiniLM-L12-v2`.
[4]`bert-base-multilingual-cased`.

| Model | English Test Set | | | Italian Test Set | | |
|---|---|---|---|---|---|---|
| | Macro | OBJ | SUBJ | Macro | OBJ | SUBJ |
| *monolingual* | en→en | | | it→it | | |
| MAJ-B | 0.33 | 0.65 | 0.00 | 0.42 | 0.85 | 0.00 |
| RND-B | 0.50 | 0.49 | 0.50 | 0.47 | 0.58 | 0.36 |
| SVM | 0.44 | 0.64 | 0.24 | 0.59 | 0.58 | 0.34 |
| LR | 0.55 | 0.63 | 0.48 | 0.60 | 0.77 | 0.42 |
| M-SBERT | 0.69 | 0.70 | 0.69 | 0.69 | 0.82 | 0.56 |
| M-BERT | 0.75 | 0.77 | 0.71 | 0.74 | **0.88** | 0.59 |
| *multilingual* | en+it→en | | | en+it→it | | |
| SVM | 0.49 | 0.63 | 0.34 | 0.60 | 0.85 | 0.35 |
| LR | 0.64 | 0.63 | 0.65 | 0.61 | 0.81 | 0.42 |
| M-SBERT | 0.71 | 0.67 | 0.76 | 0.69 | 0.81 | 0.56 |
| M-BERT | **0.80** | **0.81** | **0.80** | 0.77 | **0.88** | **0.66** |
| *crosslingual* | it→en | | | en→it | | |
| M-SBERT | 0.67 | 0.61 | 0.74 | 0.66 | 0.83 | 0.49 |
| M-BERT | 0.60 | 0.72 | 0.46 | 0.65 | 0.85 | 0.46 |

Table 4: Macro and per-class F1-score on our corpus (English) and News-ITA (Italian) test sets.

These results indicate that the multilingual setting is beneficial for certain models. Most importantly, no performance loss is observed compared to the monolingual setting, suggesting that the two corpora are coherent in their annotation. Lastly, we observe a significant performance loss for M-BERT compared to monolingual in cross-lingual. In contrast, M-SBERT achieves comparable performance in both corpora. This result is in accordance with the results of Reimers and Gurevych (2020).

## 6. Related Work

Previous work has coupled SD with sentiment analysis (Stepinski and Mittal, 2007; Chaturvedi et al., 2018), bias detection (Aleksandrova et al., 2019; Hube and Fetahu, 2019), claim extraction (Riloff and Wiebe, 2003; Banea et al., 2014), and fact-checking (Vieira et al., 2020; Jerônimo et al., 2019; Antici et al., 2021). Among the explored domains there are reviews (Pang and Lee, 2004b; Benamara et al., 2011), social media content (Volkova et al., 2013; Bosco et al., 2013), and, most notably, news media (Wiebe et al., 1999b; Antici et al., 2021; Vargas et al., 2023). While news media are generally expected to be objective, several studies observed that they can contain a substantial amount of subjective content (Riloff and Wiebe, 2003; Wahl-Jorgensen, 2013; Chong, 2019), which can be considered as an indicator of biased sentences (Vargas et al., 2023; Färber et al., 2020).

SD has been investigated at several granularity levels: at sentence level (Riloff and Wiebe, 2003;

276

Rustamov et al., 2013), segment level (Benamara et al., 2011) and document level (Antici et al., 2021). Following Vieira et al. (2020), which state that "the fragmentation of news allows the identification of subjectivity markers that cannot be identified when considering the entire documents", we build our corpus by annotating subjectivity at the sentence level.

Wiebe et al. (1999b) conducted a seminal study on subjectivity. Their work is based on *evidentiality*, which emphasizes the source of information and the primary intention of a sentence. The authors define a sentence as Objective if its primary intention is the objective presentation of facts, and as Subjective if it conveys speaker evaluations, opinions, emotions, and speculations. In contrast, we base our work on the *prescriptive paradigm* (Rottger et al., 2022): our annotation guidelines are defined to detect subjective expressions and cases relying on in-text evidence and linguistic markers, such as figures of speech, rhetorical questions, and punctuation. For example, while Wiebe et al. consider sentences that contain personal feelings as subjective because the author wants to express their subjective perspective, we consider them objective because the author can be considered a reliable and objective source of knowledge for their own feelings.

To address the inherent ambiguity of the task, Wiebe et al. enrich the binary label with a certainty rating from 0 to 3. Savinova and Moscoso Del Prado (2023) instead, define subjectivity as a range of values, hence addressing SD as a regression task. In our approach, we let annotators debate over disagreements and flag the cases in which an agreement is ultimately not reached.

Few contributions have addressed languages other than English, such as French (Benamara et al., 2011), Italian (Antici et al., 2021; Esuli and Sebastiani, 2006), and Persian (Amini et al., 2019). Banea et al. investigated SD on 6 languages using machine translation and found that multilingual information can be beneficial for SD. Similarly, Banea et al. addressed multilingual cross-lingual SD in Romanian and English documents by aligning the corresponding version of WordNet (Fellbaum, 2010). To the best of our knowledge, we are the first to address multilingual sentence-level SD experimenting over two corpora that have been manually annotated without the use of machine translation.

Our guidelines were used for the CheckThat! shared task (Galassi et al., 2023; Barrón-Cedeño et al., 2023a,b) to annotate additional data in six different languages. Part of our corpus was used in the same task as training data.

## 7. Conclusion

We present a novel set of annotation guidelines for subjectivity detection that may be applied to any language, following the prescriptive approach. We use our guidelines to create a corpus of British news articles annotated for SD at the sentence level.

We showcase the utility of the corpus by experimenting in mono-, multi-, and cross-lingual scenarios considering English and Italian. The best performance is obtained when training on articles in both English and Italian, suggesting that our methodology can be consistently applied to multiple languages.

Future research directions include evaluating our data collection methodology in other languages, domains, and tasks, such as claim verification, where subjectivity detection is relevant.

## Ethics, Limitations, and Risks

**Annotation Bias.** The perception of subjectivity is subjective in itself. It may derive from different interpretations of the language, different background knowledge, and personal biases (Chaturvedi et al., 2018). While different genders, nationalities, and areas of expertise are represented among the annotators, it is impossible to exclude that they may all share similar biases that influence their perception of subjectivity.

**Data Selection.** Some of the articles we use, especially those with many subjective sentences, may address some topics more frequently than others. Therefore, we cannot exclude the possibility that a model may learn a bias toward certain topics and the keywords associated with them. Similar problems have emerged in other NLP tasks, such as abusive language detection (Wiegand et al., 2019; Park et al., 2018). We remark that our corpus should be used only for research purposes.

**Corpus Size.** The limited size of our dataset is symptomatic of the challenging scenario we address, where data quality is of great importance and collecting a lot of high-quality data is expensive. Our annotation methodology aims at reducing annotators' bias, while producing high-quality annotations.

**Adaptation to Other Languages.** Challenges in adapting the guidelines to linguistically diverse contexts concern cultural influences reflected in a language, prior knowledge of domain-specific topics, sarcasm, and implicit information. All these factors may affect how an annotator perceives the content conveyed in the text to annotate. To address these challenges we adopt a systematic annotation methodology. In particular, our guidelines cover a set of specific abstract cases, such as sarcasm and discrimination, that are grounded in languages through the use of examples. In most cases, we believe that adapting the guidelines to new languages would require only changing these examples. A more substantial adaptation of the guidelines would be necessary if a language does not contain one of our identified cases.

**Use of LLMs.** We excluded LLM-based methods such as GPT from our work on account of the robustness and soundness of the experiments. Making a fair comparison between LLMs and classification models like ours is non-trivial due to their being highly dependent on the formulation of the prompt and probabilistic nature. Indeed, slight variations in the prompt may lead to very different results, and even the same query may produce different outputs when run multiple times (Lu et al., 2022; Sclar et al., 2023; Gan and Mori, 2023; Salinas and Morstatter, 2024).

**Impact.** Subjectivity, being an indicator of bias in the author's statement, is of valuable importance in analyzing and recognizing opinionated content. Our work could provide useful insight when studying relevant problems like fact-checking in foreign languages, where it is more difficult for researchers to recognize peculiar patterns that may hide the author's point of view. Our work aims to improve awareness and, consequently, the fairness of textual resources.

## Data and Code Availability

Our corpus, the re-annotated Italian corpus NewsSD-Ita, our guidelines, and our code are available at `https://github.com/lt-nlp-lab-unibo/newssd-eng`.

## Acknowledgements

The work of A. Galassi is supported by the European Commission NextGeneration EU programme, PNRR-M4C2-Investimento 1.3, PE00000013-"FAIR" Spoke 8. The work of F. Ruggeri is supported by the European Union's Horizon Europe research and innovation programme under GA 101070000. K. Korre's research is carried out under the project "RACHS: Rilevazione e Analisi Computazionale dell'Hate Speech in rete", in the framework of the PON programme FSE REACT-EU, Ref. DOT1303118. The work of A. Fedotova is supported by the NextGeneration EU programme, ALMArie CURIE 2021 - Linea SUpER, Ref. CUPJ45F21001470005.

## Bibliographical References

# Appendices

## A. Reproducibility

### A.1. Computing Infrastructure

We conducted all of our experiments on a machine equipped with an Intel i7 3.7GHz processor, 32 GB of RAM, and an NVidia 1080ti 11 GB.

### A.2. Model implementation

For SVM and LR we use the default configuration of `scikit-learn`. As M-SBERT we use `paraphrase-multilingual-MiniLM-L12-v2`.[5] As M-BERT we use `bert-base-multilingual-cased`.[6]

### A.3. Hyperparameters and Runtime

The models are trained with their default configuration setup. For the M-BERT model, we iterated over the training set for 4 epochs, using **Adam** as optimizer with a learning rate of $1e-5$. The model, which has around 170 million parameters, takes an average of 30 seconds per epoch during training while having an inference time of 2 seconds.

### A.4. Seeds

We perform multiple runs for the experiments. The tables report results averaged over three seed runs: 42, 1000, and 500.

### A.5. Validation Performances

Table 5 and Table 6 report classification performance on the English and Italian validation sets, respectively.

M-BERT and M-SBERT perform better than the SVM and LR baseline in the monolingual setting.

In the multilingual setting, the two models obtain contrastive results in the two languages. M-SBERT achieves the best result for English, While M-BERT for Italian.

The cross-lingual setting in English is beneficial for M-SBERT with respect to the monolingual case. Instead, The performance of M-BERT in Italian is the same in mono-lingual and cross-lingual settings.

| Model | F1-Macro | F1-OBJ | F1-SUBJ |
|---|---|---|---|
| *mono-lingual: en → en* | | | |
| SVM | 0.50 | 0.63 | 0.36 |
| LR | 0.56 | 0.62 | 0.49 |
| M-SBERT | 0.63 | 0.63 | 0.62 |
| M-BERT | 0.62 | 0.67 | 0.57 |
| *multi-lingual: en+it → en* | | | |
| SVM | 0.55 | **0.77** | 0.32 |
| LR | 0.60 | 0.72 | 0.47 |
| M-SBERT | **0.71** | 0.69 | **0.74** |
| M-BERT | 0.61 | 0.66 | 0.56 |
| *cross-lingual: it → en* | | | |
| M-SBERT | 0.66 | 0.62 | 0.70 |
| M-BERT | 0.41 | 0.64 | 0.19 |

Table 5: Classification performance on our corpus validation set.

| Model | F1-Macro | F1-OBJ | F1-SUBJ |
|---|---|---|---|
| *mono-lingual: it → it* | | | |
| SVM | 0.51 | 0.82 | 0.21 |
| LR | 0.57 | 0.78 | 0.35 |
| M-SBERT | 0.73 | 0.82 | **0.64** |
| M-BERT | 0.71 | **0.87** | 0.55 |
| *multi-lingual: en+it → it* | | | |
| SVM | 0.55 | 0.77 | 0.32 |
| LR | 0.60 | 0.72 | 0.47 |
| M-SBERT | 0.60 | 0.72 | 0.47 |
| M-BERT | **0.74** | **0.87** | 0.62 |
| *cross-lingual: en → it* | | | |
| M-SBERT | 0.70 | 0.84 | 0.57 |
| M-BERT | 0.71 | **0.87** | 0.55 |

Table 6: Classification performance on SubjectivITA validation set.

## B. SubjectivITA and NewsSD-ITA

Among the main differences between our and SubjectivITA's annotation guidelines, is the labeling of quotes and explicit emotions. With the purpose of increasing the compatibility between the two corpora, we partially re-annotate SubjectivITA, labeling sentences that belong to these categories as objective. We refer to the corpus thus created as NewsSD-ITA. The final distribution of classes in the NewsSD-ITA corpus is presented in Table 7.

|  | # Art. | # Sent. | # OBJ | # SUBJ |
|---|---|---|---|---|
| Train | 80 | 1,399 | 1,079 | 320 |
| Val | 13 | 214 | 152 | 62 |
| Test | 10 | 227 | 167 | 60 |
| Total | 103 | 1,840 | 1,398 | 442 |

Table 7: Statistics of NewsSD-ITA corpus.

---

[5] https://huggingface.co/sentence-transformers/paraphrase-multilingual-MiniLM-L12-v2
[6] https://huggingface.co/google-bert/bert-base-multilingual-cased

## C. Annotation Guidelines

**Definitions**

**Subjective.** A sentence is considered **subjective** when it is based on —or influenced by— personal feelings, tastes, or opinions. Otherwise, the sentence is considered **objective**.

Table 8 shows an example for each type. Sentence 1 is **objective** because the author describes a historical event without giving any opinion or personal comment. In contrast, in Sentence 2 the author explicitly conveys their personal emotions, making the sentence subjective.

**Specific Subjective Cases**

**SUBJ 1.** A sentence is subjective if it **explicitly** reports the **personal opinion** of its author. Rhetorical questions are considered as an expression of an opinion as well; see Ex. (c). Additionally, speculations which draw conclusions are considered as opinions, see Ex. (d). Examples:

(a) **It has everything you could want in a holiday**: **beautiful** sandy beaches and clear waters, ancient history and culture, **delicious** food (**the Greek salads are simply on another level**), island hopping, nightlife and more.

(b) After treading vineyard soils and seeing grapes ripening, **that merlot becomes more than just a Wednesday night relaxant**.

(c) Do they really think other nations sprouted up out of the ground?

(d) But Putin **will hope to sow uncertainty** in the eyes of policymakers' meetings in New York.

**SUBJ 2.** A sentence is subjective if it contains **sarcastic** or **ironic** expressions attributable to its author.
Examples:

(e) It's no lie that the USA is **one heck of a big country (said in a southern twang)**.

(f) With Land Rover bowdlerising images of the vehicle into **little more than a perfume advertisement** on TV[...].

(g) Especially if you're more excited at the prospect of sampling rare bottles from the cellar **than snapping vineyard selfies**.

**SUBJ 3.** A sentence is subjective if it contains **exhortations** of **personal auspices** made by its author.
Examples:

(h) **The West should arm Ukraine faster**.

**SUBJ 4.** A sentence is subjective if it contains **discriminating** or **downgrading** expressions.
Examples:

(i) And what is even more evident is the **perverse** role reversal that is taking place, in which he who sits in Rome has the task of formulating heterodox principles opposed to Catholic doctrine, and his accomplices in the Dioceses have the role of **scandalously** applying them, in an **infernal** attempt to undermine the Moral law in order to obey the spirit of the world.

(j) How did we reach the stage where priests and bishops **cowered like frightened puppies** before a common flu, where their predecessors ministered fearlessly among the lepers, the cripples, and the victims of typhoid, cholera, smallpox, and Bubonic Plague?

**SUBJ 5.** A sentence is subjective if it contains rhetorical figures, like **hyperboles**, explicitly made by its author to convey their opinion.
Examples:

(k) Barcelona where it all began, Messi was a **king** in Catalonia and he lived like one too.

(l) The churches, and the Catholic Church in particular (which is by far the largest), had the ability to put an end to the lockdown **madness** and the **COVID-terror campaign**, had they wished to do so.

(m) So it must be biochemistry that is really **what is racist**.

**Specific Cases of Objectivity**

If a sentence does not meet any subjectivity type listed in the previous section, it is considered objective. Here we include examples of **objective** sentences which may be wrongly interpreted as subjective.

**Case 1.** A sentence is objective when it reports on **news** or **historical facts** that are quoted by the author of the sentence.
Examples:

(a) President Putin has just reiterated his threat to use nuclear weapons and announced that Russian-controlled Ukrainian territory will become part of the Russian Federation.

(b) In the modern era electroconvulsive therapy, **first used in 1938**, **became a treatment for some serious forms of depression in the post-war decades**.

**Case 2.** A sentence is objective when it describes the **personal feelings, emotions or moods of the writer**, without conveying opinions on other matters.
Examples:

(c) I was definitely **surprised** at how emotional **I felt** watching the service.

(d) The second I saw him, **I felt a jolt of connection**.

| | Sentence | Label |
|---|---|---|
| 1. | India, who was the bridesmaid at the King's wedding to Princess Diana in 1981, could not be seen in the footage, but filmed the video as she walked through the grounds of the royal residence. | OBJ |
| 2. | It is a sad truth that many of the villages in this region of Portugal are dying. | SUBJ |

Table 8: Examples of subjective and objective sentences.

**Case 3.** A sentence is objective if it expresses an opinion, claim, emotion or a point of view that is **explicitly attributable to a third-party** (e.g., a person mentioned in the text).
Examples:

(e) **Frank Drake believed** that the universe had to contain other intelligent beings.

(f) "You showed callous indifference to Dean's fate after he had been repeatedly stabbed" **the judge said.**

**Note**: The presence of quotation marks (" "), when used to quote a third person (be it at the beginning of the sentence, at the end, or both), represents an explicit third-party opinion, even if it is not clearly stated in the sentence.
Examples:

(g) **"**Crosbie is an extremely violent man who has no place in society, and we welcome the jury's verdict today.**"**

(h) **"**My children have lost their hero and I have lost my chosen person - the person I chose to spend my life with.

(i) For these reasons and out of conviction, I consider myself bound in my conscience to say no.**"**

**Case 4.** A sentence is objective if it contains a **comment** made by the author of the sentence that **does not draw any conclusion**. In particular, the author doesn't convey their personal interpretation or opinion, leaving the discussion on the topics of interest open.
Examples:

(j) **It is not clear yet which of the couples** from the E4 reality show remain together and who have now, **because the series has not concluded**.

(k) Do car manufacturers know how far their EVs will really go?

(l) Exact figures **are hard to come by**, but Ukraine **may** well have more troops available than Russia now.

**Case 5.** A sentence is objective if it contains **factual conclusions** made by the author of the sentence that **do not convey any stance or personal opinion**, or are justified up by a non-personal hypothesis.
Examples:

(m) In years gone by, travel to Japan was notoriously expensive, but **the devaluing of the yen has made it more accessible**.

(n) The bottom-up approaches which target the molecular, genetic and electrical fundamentals of the brain **can assist top-down approaches to brain disorder such as talking therapies**.

(o) **Based on our experiences and road tests**, a good rule of thumb is to expect to achieve somewhere between 75 and 80 per cent of a car's WLTP Combined range[...]

**Case 6.** When referring to an individual, any kind of **well-known nickname** or **title** is considered objective.
Examples:

(p) Things have certainly progressed on the pitch for **Spurs** this season.

(q) The **Duke of York** 'plotted' with Diana to 'push **Prince** Charles aside'.

**Case 7.** Any kind of **common expression** or **proverb** is considered objective.
Examples:

(r) the adage **'sticks and stones may break my bones, but words can never hurt me'**.

(s) **Home sweet home**: George poses in one of the rooms at his sprawling Hampstead home during a photoshoot in 2002.